\documentclass{article}

\usepackage[preprint]{colm2025_conference}

\usepackage{microtype}
\usepackage{hyperref}
\usepackage{url}
\usepackage{booktabs}
\usepackage{multirow}

\usepackage[ruled, linesnumbered]{algorithm2e}
\usepackage{amsmath}

\usepackage{amsmath}
\usepackage{amsthm}
\usepackage{amssymb}
\usepackage{amsfonts}
\usepackage{dsfont}
\usepackage{graphicx}
\usepackage{adjustbox}
\usepackage{caption}
\usepackage{subcaption}

\usepackage{booktabs}
\usepackage{multirow}
\usepackage{siunitx}
\usepackage{lineno}

\definecolor{darkblue}{rgb}{0, 0, 0.5}
\hypersetup{colorlinks=true, citecolor=darkblue, linkcolor=darkblue, urlcolor=darkblue}

\theoremstyle{plain}
\newtheorem{theorem}{Theorem}

\newtheorem{proposition}[theorem]{Proposition}
\newtheorem*{proposition*}{Proposition}

\theoremstyle{definition}

\newtheorem*{definition*}{Definition}

\theoremstyle{remark}

\title{Fail Fast, or Ask: Mitigating the Deficiencies of Reasoning LLMs with Human-in-the-Loop Systems Engineering}

\author{Michael J. Zellinger \\ \textbf{Matt Thomson}\\
California Institute of Technology\\
\texttt{\{zellinger, mthomson\}@caltech.edu} \\
}

\date{}

\RequirePackage{natbib}
\begin{document}

\ifcolmsubmission
\linenumbers
\fi

\maketitle

\begin{abstract}
State-of-the-art reasoning LLMs are powerful problem solvers, but they still occasionally make mistakes. However, adopting AI models in risk-sensitive domains often requires error rates near 0\%. To address this gap, we propose collaboration between a reasoning model and a human expert who resolves queries the model cannot confidently answer. We find that quantifying the uncertainty of a reasoning model through the length of its reasoning trace yields an effective basis for deferral to a human, e.g., cutting the error rate of Qwen3 235B-A22B on difficult MATH problems from 3\% to less than 1\% when deferring 7.5\% of queries. However, the high latency of reasoning models still makes them challenging to deploy on use cases with high query volume. To address this challenge, we explore fronting a reasoning model with a large non-reasoning model. We call this modified human-in-the-loop system ``Fail Fast, or Ask'', since the non-reasoning model may defer difficult queries to the human expert directly (``failing fast''), without incurring the reasoning model's higher latency. We show that this approach yields $\approx 40\%$ latency reduction and $\approx 50\%$ cost savings for DeepSeek R1 while maintaining 90+\% area under the accuracy-rejection curve. However, we observe that latency savings are lower than expected because of \textit{latency drag}—the phenomenon that processing easier queries with a non-reasoning model pushes the reasoning model's latency distribution towards longer latencies. Broadly, our results suggest that the deficiencies of state-of-the-art reasoning models—nontrivial error rates and high latency—can be substantially mitigated through black-box systems engineering, without requiring access to LLM internals.
\end{abstract}

\maketitle

\section{Introduction}

Reasoning LLMs are trained using reinforcement learning to ``think'' until finding the correct answer to a query (\citealp{deepseekai2025}). This approach has significantly lowered error rates on complex tasks (\citealp{openai2024}; \citealp{yang2025}), making ``thinking'' a dominant paradigm for state-of-the-art language models.

However, many risk-sensitive domains expect error rates near 0\%—a high bar which even state-of-the-art reasoning models may fail to meet.

Moreover, reasoning models suffer from high latency (\citealp{zellinger2025econ}), making these models unsuitable for tasks where answers must be delivered quickly. Query response times measured in seconds and even minutes can arise for difficult tasks, making standard user interaction patterns—which are based on rapid feedback—problematic (\citealp{kohavi2007}). A similar challenge arises for batch workloads involving a large volume of queries. For example, with a per-query latency of 30 seconds—not unusual for reasoning models—crunching through a million database records takes around 1 year.

In this paper, we explore a human-in-the-loop systems approach (``Fail Fast, or Ask'') for mitigating these deficiencies of reasoning models, with the goal of speeding their path towards practical deployment on real-world tasks.

\begin{figure}
    \includegraphics[width=\textwidth]{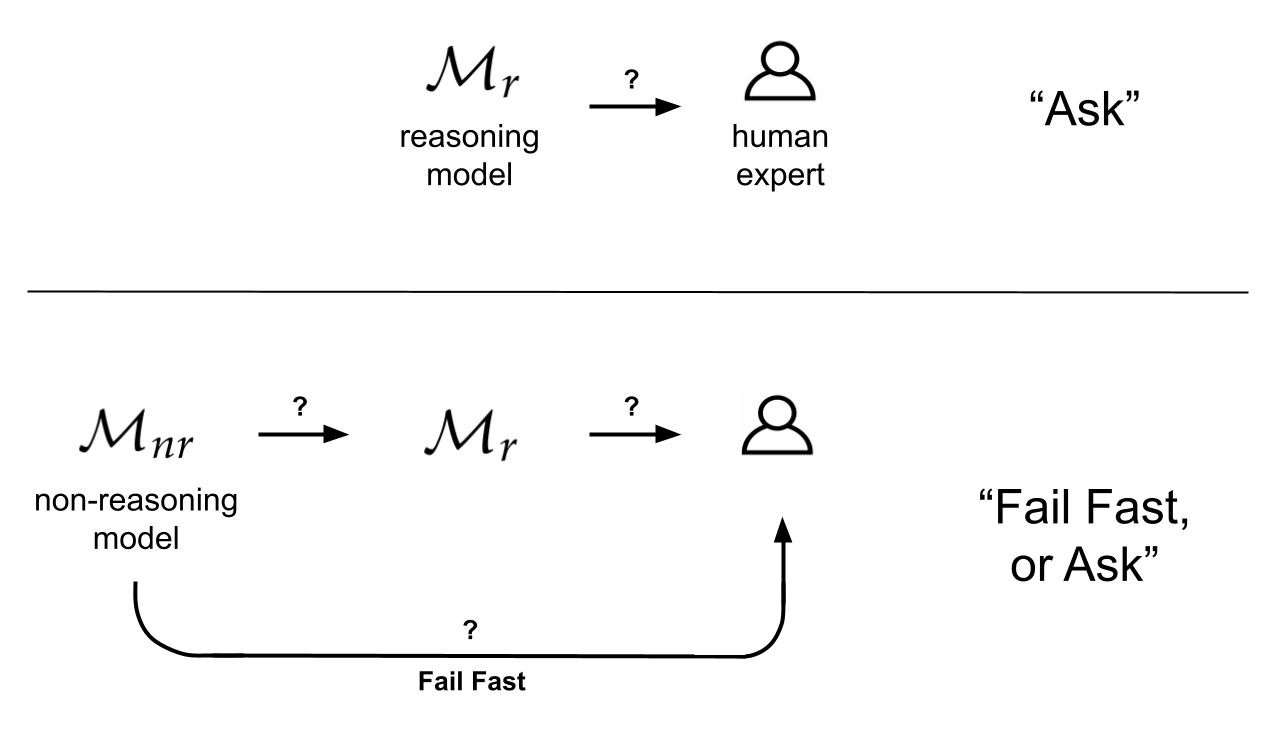}
    \caption{We explore two systems aimed at reducing deficiencies of reasoning models. Above: ''Ask'' aims to reduce the error rate of reasoning model $\mathcal{M}_r$ by deferring difficult queries to a human expert when $\mathcal{M}_r$ is uncertain. Bottom: ``Fail Fast, or Ask'' aims to offset the reasoning model's high latency by fronting it with a faster non-reasoning model $\mathcal{M}_{nr}$, which may directly defer queries to the human expert (``failing fast'').}
    \label{fig:fail_fast_or_ask_diagram}
\end{figure}

Described simply, our ``Fail Fast, or Ask'' system applies distinct strategies for improving 1) the error rate and 2) the latency of a reasoning model $\mathcal{M}_r$.

To achieve error reduction, $\mathcal{M}_r$ defers difficult queries to a human expert $\mathcal{H}$. Similar to prior successes on achieving near-zero error rates on image classification (\citealp{geifman2017}), this strategy lowers the error rate of $\mathcal{M}_r$ at the expense of abstaining from a nonzero fraction of queries.

Second, to speed up response times we send each query first to a faster, non-reasoning model $\mathcal{M}_\text{nr}$. This model may respond directly to the query, defer it to $\mathcal{M}_\text{nr}$, or send it to the human expert $\mathcal{H}$.

Our experiments yield the following conclusions:
\begin{enumerate}
    \item Quantifying the uncertainty of reasoning models by the lengths of their thinking traces effectively improves accuracy for DeepSeek R1 and Qwen3 235B-A22B, but not OpenAI o3.
    \item Deferring difficult queries to a human expert achieves +2\% accuracy, pushing the accuracy of Qwen3 235B-A22B on the most challenging MATH questions from 97\% to 99+\%.
    \item Our ``Fail Fast, or Ask'' system, which fronts a slow reasoning model with a faster non-reasoning model (Llama3.1 405B), maintains 90+\% accuracy while responding to queries $\approx 40\%$ faster.
\end{enumerate}

\section{Related Work}

\noindent \textbf{Quantifying the Uncertainty of Reasoning Models}. Previous work suggests that longer reasoning traces correlate with lower accuracy (\citealp{ballon2025}), resulting from higher perceived problem difficulty (\citealp{shojaee2025}; \citealp{lawsen2025}).

Interestingly, this phenomenon highlights nuances of causation vs correlation. While longer reasoning traces \textit{correlate} with higher error rates, deliberately using a greater number of reasoning tokens paradoxically \textit{causes} improved performance (\citealp{jin2024}; \citealp{muennighoff2025}). Such apparent paradoxes are well-explained in the causal inference literature (\citealp{pearl2009}) and are due to adverse selection effects. Specifically, when \textit{forcing} a model to use a greater number of reasoning tokens, we condition on all queries, whereas if we \textit{observe} a model using a greater number of reasoning tokens, we condition on a particularly difficult set of queries.\footnote{This scenario closely mirrors the common situation in hospitals where only the sickest patients are assigned the strongest medication. Naive statistical analysis then leads to the erroneous conclusion that the medication lowers patients' survival rate.}

\noindent \textbf{Selective Prediction with Large Language Models}. Selective prediction (\citealp{elyaniv2010}) gives a machine learning model the ability to abstain from answering difficult queries, making very low error rates possible. Using this approach, \citet{geifman2017}) have demonstrated near-zero error rates on top-5 ImageNet classification. In natural language processing (NLP), selective prediction has not been as popular, except for early contributions (\citealp{xin2021}; \citealp{varshney2022}; \citealp{ren2023}). Once NLP entered the LLM era, several authors have proposed uncertainty metrics for predicting LLM errors (\citealp{manakul2023}; \citealp{azaria2023}; \citealp{chen2023}; \citealp{farquhar2024}). These contributions typically evaluate the quality of their proposed uncertainty metrics by simulating a selective prediction task in which the LLM only answers queries for which its uncertainty is low. This evaluation results in a plot of conditional accuracy vs rejection rate, and the area under this curve (AUARC) quantifies performance across rejection rates.

\noindent \textbf{Human-in-the-Loop AI Systems}. Collaboration between AI systems and humans has been an active research topic in human-computer interaction (\citealp{amershi2024}; \citealp{kamar2016}; \citealp{shneiderman2020}; \citealp{watkins2025}), with heightened interest in application domains such as medicine. Some of this work has explored the complementary strengths of human experts and AI models (\citealp{dvijotham2023}), while others treat human experts as an omniscient oracle—a safe fallback for imperfect AI models (\citealp{strong2025}; \citealp{fanconi2025}). Our work takes this latter approach, although we view explicit modeling of human errors as a promising avenue for follow-up research.

\section{Fail Fast, or Ask}
\label{sec:core}

We consider two human-in-the-loop systems, as illustrated in Figure \ref{fig:fail_fast_or_ask_diagram}. The first system (``Ask'') consists of a large reasoning model $\mathcal{M}_r$ that defers difficult queries to a human expert $\mathcal{H}$. The second system (``Fail Fast, or Ask'') additionally contains a large non-reasoning model $\mathcal{M}_\text{nr}$, which fields all queries; based on its confidence, $\mathcal{M}_\text{nr}$ either 1) responds to the query, 2) passes the query to the reasoning model $\mathcal{M}_\text{r}$, or 3) directly defers the query to the human expert (``failing fast'').

\noindent \textbf{Predicting Reasoning Model Errors}. Prior work has revealed that reasoning models emit more thinking tokens when they are uncertain about the answer to a query (\citealp{ballon2025}). Leveraging this insight, we predict reasoning model errors by simply thresholding the number of output tokens. For example, to reject 10\% of queries (and defer them to the human expert), we set the output token threshold at the 90\% quantile of its empirical distribution.

\noindent \textbf{Predicting Non-Reasoning Model Errors}. To predict the errors of $\mathcal{M}_\text{nr}$, we estimate its confidence using the P(True) strategy of \citet{kadavath2022}. Prior work shows that this methodology yields good performance for Llama3.1 405B (\citealp{zellinger2025econ}).

On each query, the non-reasoning model chooses between three possible actions: 1) return the response to the query (\texttt{respond}), 2) pass the query to the reasoning model (\texttt{pass}), or 3) directly defer the query to the human expert (\texttt{fail fast}). Its action $\pi_{nr}$ is based on simple thresholding of its P(True) confidence estimate $p_\text{true}$:
\begin{equation}
    \pi_{nr} = \begin{cases}
                  \texttt{fail fast}      & \text{if } p_\text{true} \leq c_\text{fail fast}, \\[6pt]
                  \texttt{pass}        & \text{if } p_\text{true} > c_\text{fail fast} \text{ and } p_\text{true} \leq c_\text{pass}, \\[6pt]
                  \texttt{respond} & \text{if } p_\text{true} > c_\text{pass}.
                \end{cases}
\end{equation}


\noindent \textbf{Configuration of the ``Fail Fast, or Ask'' System}. To modulate utilization of the faster non-reasoning model $\mathcal{M}_{nr}$, we fix its utilization rate $u := \mathbb{P}(\pi_{nr} \neq \texttt{pass})$. This rate denotes the proportion of queries processed without involving the reasoning model $\mathcal{M}_r$; in other words, $1-u$ is the rate at which $\mathcal{M}_{nr}$ passes queries to $\mathcal{M}_r$.

To set the rate $\mathbb{P}(\pi_{nr} = \texttt{fail fast})$, we start with a desired overall rejection rate $\mathbb{P}(\rightarrow \mathcal{H})$, where $\mathcal{H}$ represents the human expert. We enforce that the reasoning model's conditional rejection rate match the system's overall rejection rate:
\[
\mathbb{P}(\mathcal{M}_r \rightarrow \mathcal{H} ~|~ \mathcal{M}_{nr} \rightarrow \mathcal{M}_r) = \mathbb{P}(\rightarrow \mathcal{H}).
\]
This assumption sets the non-reasoning model's fail-fast rate at
\[
\mathbb{P}(\pi_{nr} = \texttt{fail fast}) =  \mathbb{P}(\rightarrow \mathcal{H}) - (1-u) ~ \mathbb{P}(\mathcal{M}_r \rightarrow \mathcal{H} ~|~ \mathcal{M}_{nr} \rightarrow \mathcal{M}_r).
\]
This simple methodology yields an accuracy-rejection trade-off for each choice of the system's overall rejection rate $\mathbb{P}(\text{reject})$ and the non-reasoning model's utilization rate $u$, without requiring any training or optimization. Choosing $u > 0$ yields latency and cost savings relative to the baseline ``Ask'' system consisting only of a reasoning model $\mathcal{M}_r$ and human expert (Figure \ref{fig:fail_fast_or_ask_diagram}), as Figure \ref{fig:utilization} demonstrates.

\noindent \textbf{Human Expert}. We assume that the human expert can instantly and perfectly answer each query. Thus, we treat deferral to the expert in the same manner as rejections of the query in selective prediction, and do not further model the human expert's error rate or the time he/she requires to resolve a query. However, it would be interesting to study these aspects in future work, as we note in Section \ref{sec:discussion}.

\noindent \textbf{Metrics}. We measure system performance by the conditional accuracy $A(r)$ (or error $E(r) = 1 - A(r)$) for different ``rejection rates'' $r$ of deferring queries to the human expert (\citealp{elyaniv2010}). $A(r)$ is conditioned on the non-rejected queries; this ensures that accuracy does not improve automatically by raising the rejection rate. This approach yields \textit{accuracy-rejection curves} charting $A(r)$ vs $r$. To summarize performance in a single number, we follow prior work and report the \textit{area under the accuracy-rejection curve} (AUARC). We consider rejection rates from 0\% to 20\% and report AUARC as the mean conditional accuracy across these rejection rates.

\section{Latency Drag}

It may appear that deferring queries from the non-reasoning model $\mathcal{M}_\text{nr}$ to the reasoning model $\mathcal{M}_r$ at a rate $1-u$ will yield a system latency of
\begin{equation}
    \label{eq:ideal_latency}
    \mathbb{E}[L] = u~\mathbb{E}[L_\text{nr}] + (1-u)~(\mathbb{E}[L_\text{nr}]+ \mathbb{E}[L_\text{r}]),
\end{equation}
where $L_r$ is the latency of the reasoning model and $L_\text{nr}$ is the latency of the non-reasoning model. Unfortunately, this expectation is mistaken, for a reason we term \textit{latency drag}. Predicting errors by the reasoning model relies on the empirical correlation between high uncertainty and longer reasoning traces. Precisely this same correlation implies that conditioning on difficult queries (those passed by the non-reasoning model $\mathcal{M}_\text{nr}$) results in \textit{longer latencies}.

\begin{figure}[htbp]
    \centering
    \begin{subfigure}[b]{0.45\textwidth}
        \centering
        \includegraphics[width=\textwidth]{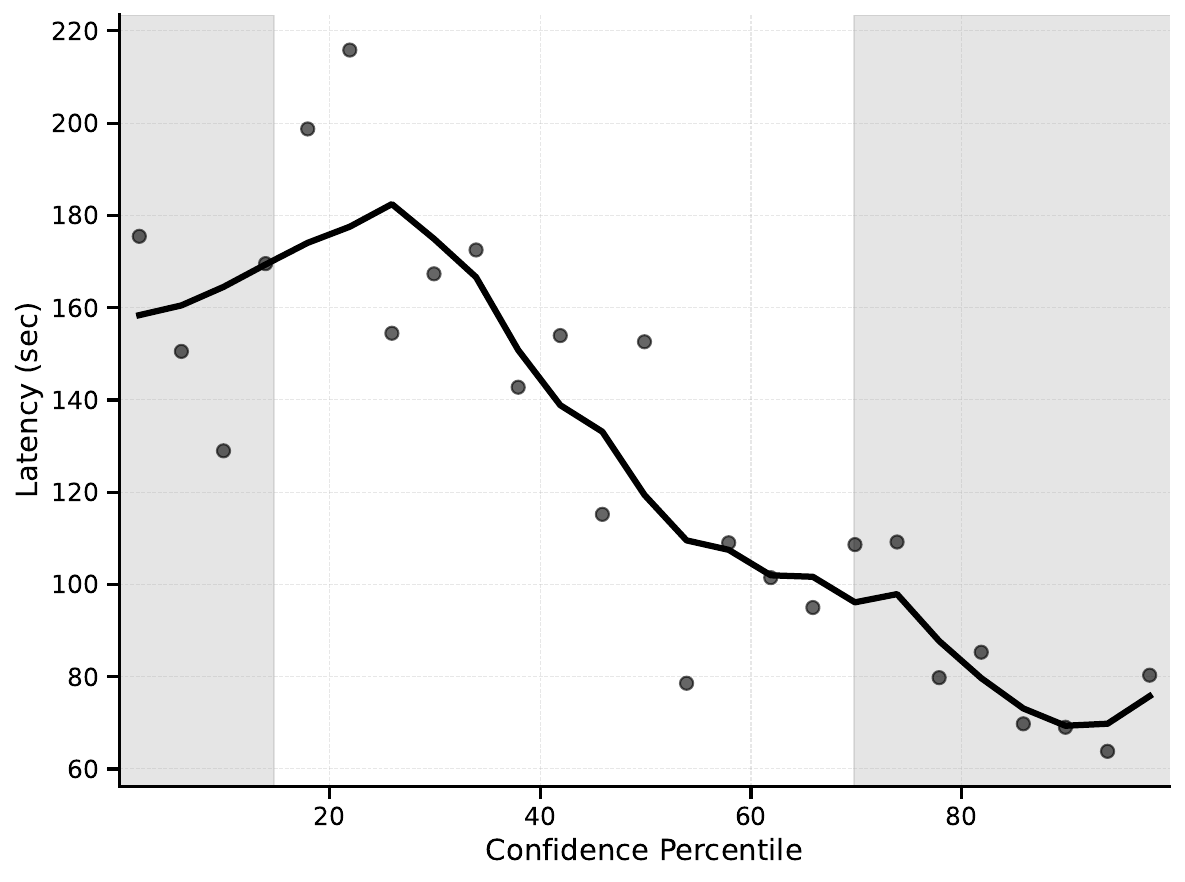}
        \caption{Qwen3 235B-A22B}
        \label{fig:latency_drag_a}
    \end{subfigure}
    \hfill
    \begin{subfigure}[b]{0.45\textwidth}
        \centering
        \includegraphics[width=\textwidth]{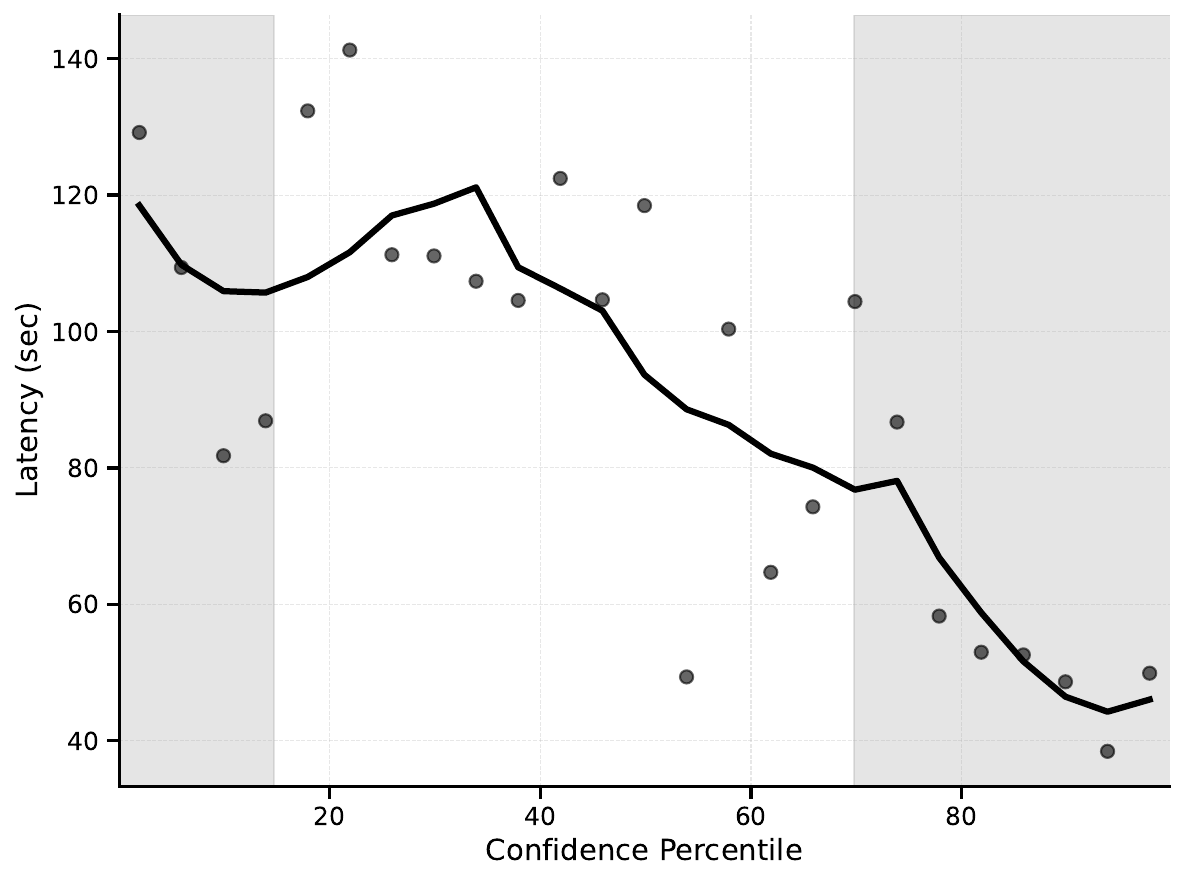}
        \caption{DeepSeek R1}
        \label{fig:latency_drag_b}
    \end{subfigure}
    \caption{Fronting the reasoning model $\mathcal{M}_r$ with a non-reasoning model $\mathcal{M}_{nr}$ increases the reasoning model's average latency (white area), leading to lower latency savings than expected. Fundamentally, this \textit{latency drag} results from the negative correlation between the reasoning model's latency ($y$ axis) and the non-reasoning model's confidence ($x$ axis).}
    \label{fig:latency_drag}
\end{figure}

Figure \ref{fig:latency_drag_a} shows the effect of latency drag for DeepSeek R1 and Qwen3 225B-A22. For lower confidence percentiles of Llama3.1 405B (\textit{x} axis), the conditional latency of the reasoning models (\textit{y} axis) rises. As a result, deferring queries from Llama3.1 405B to the reasoning model results in higher latencies than predicted by Equation (\ref{eq:ideal_latency}).

More precisely, fronting the reasoning model $\mathcal{M}_r$ with a faster non-reasoning model alters the latency distribution of $\mathcal{M}_r$ in two ways, eliminating both low \textit{and} high latencies: the former by never passing easy queries, and the latter by ``failing fast'' on the most difficult queries. The gray shaded areas in Figure \ref{fig:latency_drag} illustrates these two competing effects. In practice, latency drag from eliminating easy queries outweighs latency reduction from ``failing fast.''

\section{Selective Prediction Performance}

We evaluate the selective prediction performance of both the ``Ask'' system $\mathcal{M}_r \rightarrow \mathcal{H}$ and the ``Fail Fast, or Ask'' system $[\mathcal{M}_{nr} \rightarrow \mathcal{M}_r] \rightarrow \mathcal{H}$, for different degrees of utilizing the non-reasoning model.

To address the utility of our methodology in a complex problem-solving domain where reasoning models excel, we evaluate on 500 questions with maximum difficulty (5/5) from the MATH benchmark (\citealp{hendrycks2021}). We use the same data as \citet{zellinger2025econ}, which is filtered for questions with numeric answers.

We consider three state-of-the-art large reasoning models: Qwen3 225B-A22 (\citealp{yang2025}), DeepSeek R1 (\citealp{deepseekai2025}), and OpenAI o3. As the non-reasoning model, we use Llama3.1 405B (\citealp{grattafiori2024}) throughout, since prior work suggests it effectively quantifies its own uncertainty on MATH (\citealp{zellinger2025econ}). Table \ref{tab:baseline_performance} summarizes the baseline performance of these LLMs.

\begin{table*}[t]
\centering
\small
\caption{Baseline performance metrics for reasoning ($\mathcal{M}_r$) and non-reasoning ($\mathcal{M}_{nr}$) models for difficult MATH questions. For each metric, we report the average value per query.}
\label{tab:baseline_performance}
\resizebox{\textwidth}{!}{%
\begin{tabular}{lcccccc}
\toprule
\textbf{Model} & \textbf{Role} & \textbf{Error Rate (\%)} & \textbf{Latency (sec)} & \textbf{Cost (\$)} & \textbf{Output Tokens (\#)} \\
\midrule
qwen3-235b-a22b & $\mathcal{M}_r$ & 2.8 & 125.9 & $9.5 \times 10^{-3}$ & 10.8K \\
deepseek-r1-0528 & $\mathcal{M}_r$ & 5.8 & 89.7 & $7.9 \times 10^{-2}$ & 9.8K \\
o3-2025-04-16 & $\mathcal{M}_r$ & 3.8 & 67.6 & $1.9 \times 10^{-2}$ & 2.2K \\
llama-v3p1-405b-instruct & $\mathcal{M}_{nr}$ & 30.6 & 12.4 & $3.6 \times 10^{-3}$ & 978 \\
\bottomrule
\end{tabular}%
}
\end{table*}


\subsection{How well can we quantify the uncertainty of reasoning models?}

\begin{figure}[htbp]
    \centering
    \begin{subfigure}[b]{0.3\textwidth}
        \centering
        \includegraphics[width=\textwidth]{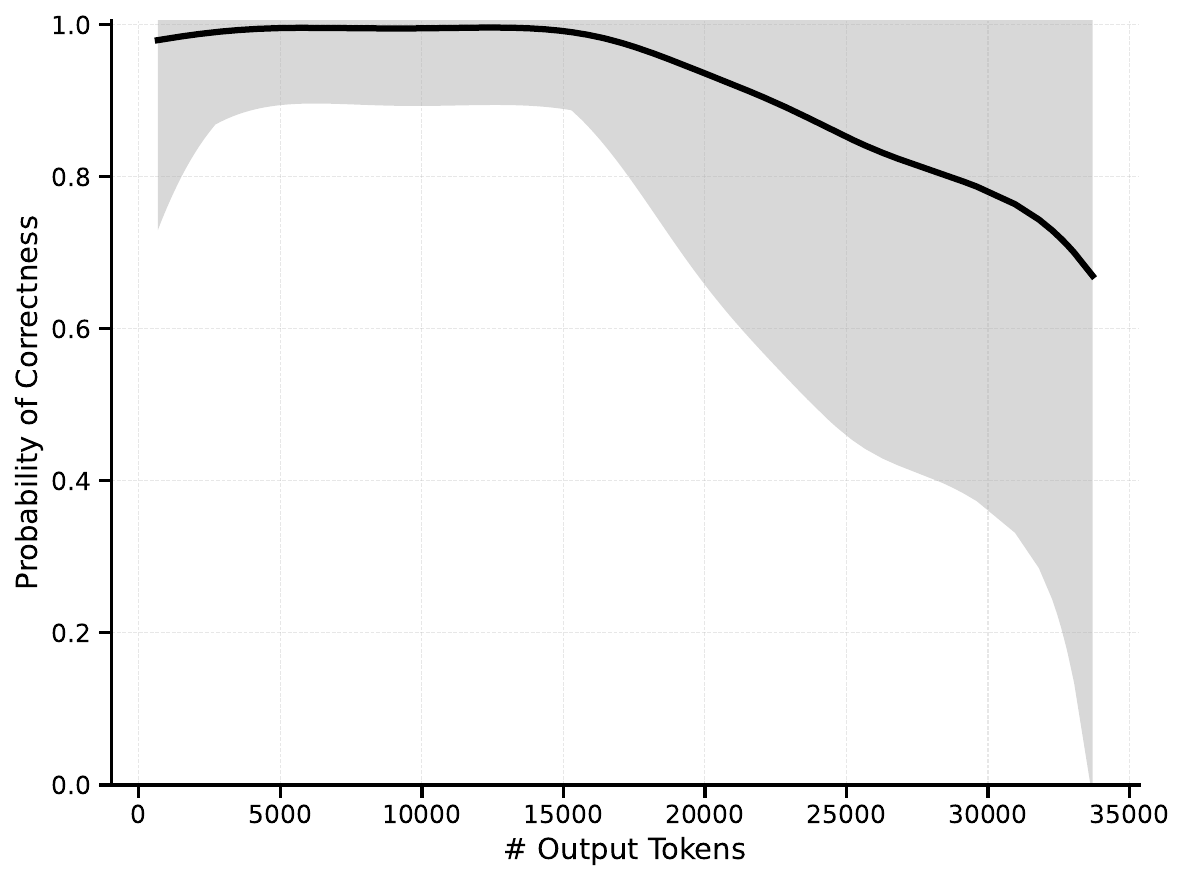}
        \caption{Qwen3 235B-A22B}
        \label{fig:uncertainty_quantification_a}
    \end{subfigure}
    \hfill
    \begin{subfigure}[b]{0.3\textwidth}
        \centering
        \includegraphics[width=\textwidth]{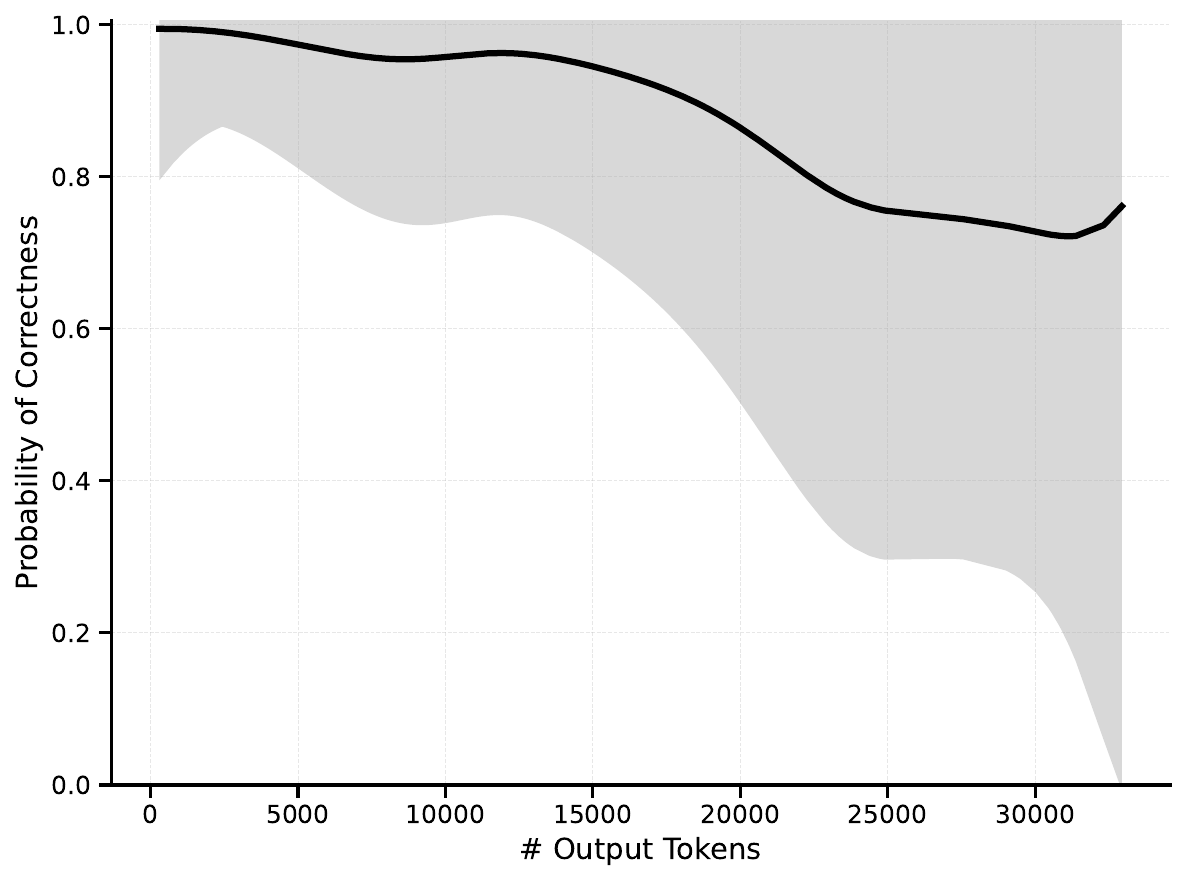}
        \caption{DeepSeek R1}
        \label{fig:uncertainty_quantification_b}
    \end{subfigure}
    \hfill
    \begin{subfigure}[b]{0.3\textwidth}
        \centering
        \includegraphics[width=\textwidth]{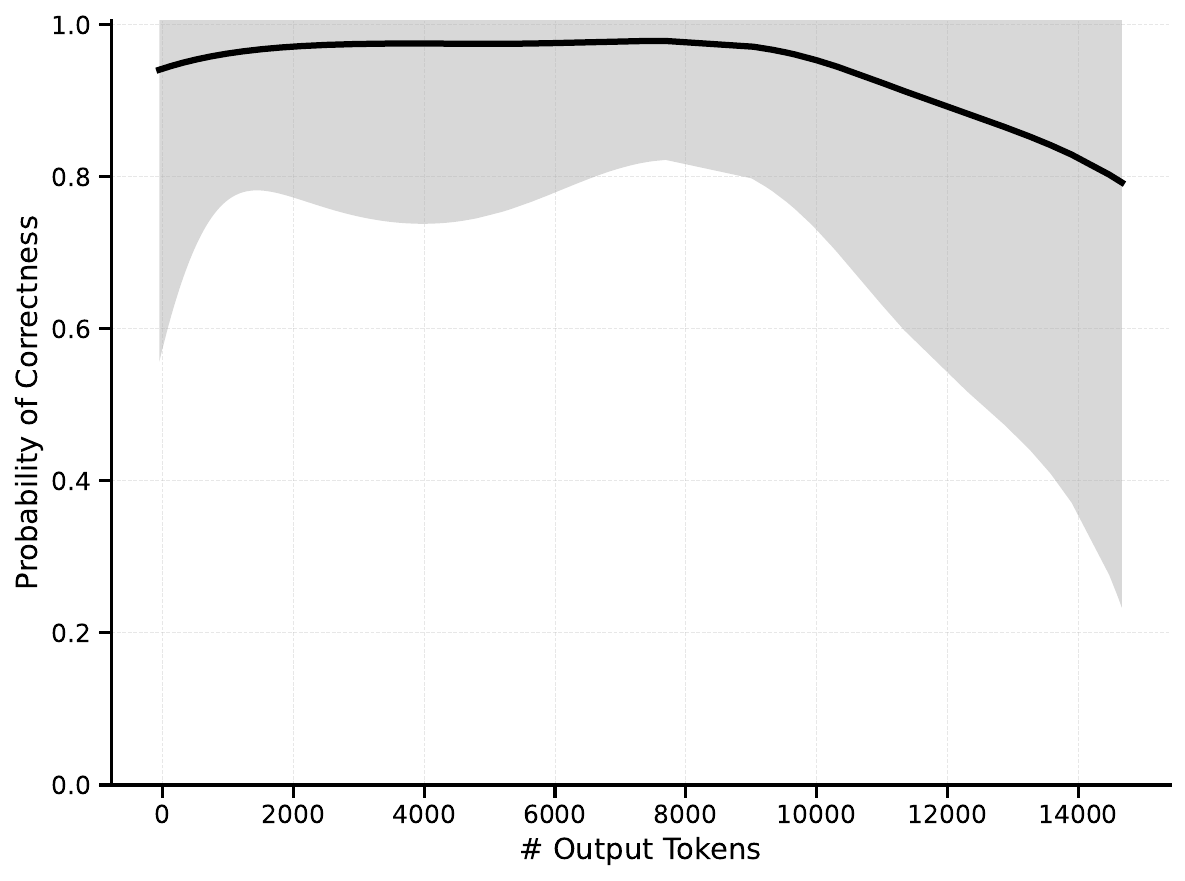}
        \caption{OpenAI o3}
        \label{fig:uncertainty_quantification_c}
    \end{subfigure}
    
    \caption{Local linear regression ($\pm1\sigma$) of the reasoning models' correctness vs the number of output tokens shows that correctness decreases when reasoning traces are long.}
    \label{fig:uncertainty_quantification}
\end{figure}

Figure \ref{fig:uncertainty_quantification} shows the results of using the length of reasoning models' thinking traces to predict their correctness on answering queries. Each plot shows a local linear regression computing the conditional expectation of correctness for a specified number of output tokens. Each curve shows decreased accuracy when the number of output tokens is large. 

However, we observe some variability between the models. Although the shape of each curve is similar—initially flat, then steeply declining—this pattern is attenuated for OpenAI o3, which displays a longer flat region and smaller decline in accuracy. 

\subsection{How low does selective prediction push the error rates of reasoning models?}

\begin{figure}[htbp]
    \centering
    \begin{subfigure}[b]{0.3\textwidth}
        \centering
        \includegraphics[width=\textwidth]{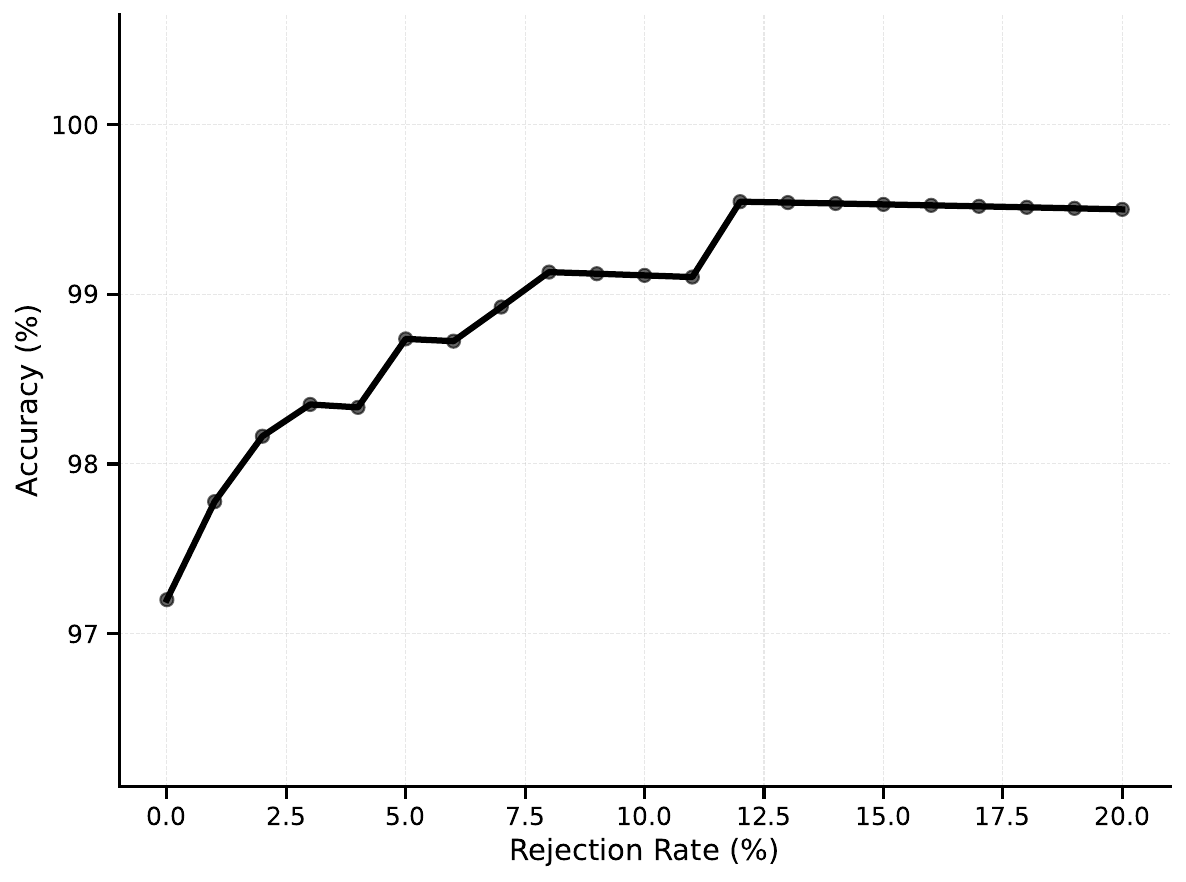}
        \caption{Qwen3 235B-A22B}
        \label{fig:selective_prediction_a}
    \end{subfigure}
    \hfill
    \begin{subfigure}[b]{0.3\textwidth}
        \centering
        \includegraphics[width=\textwidth]{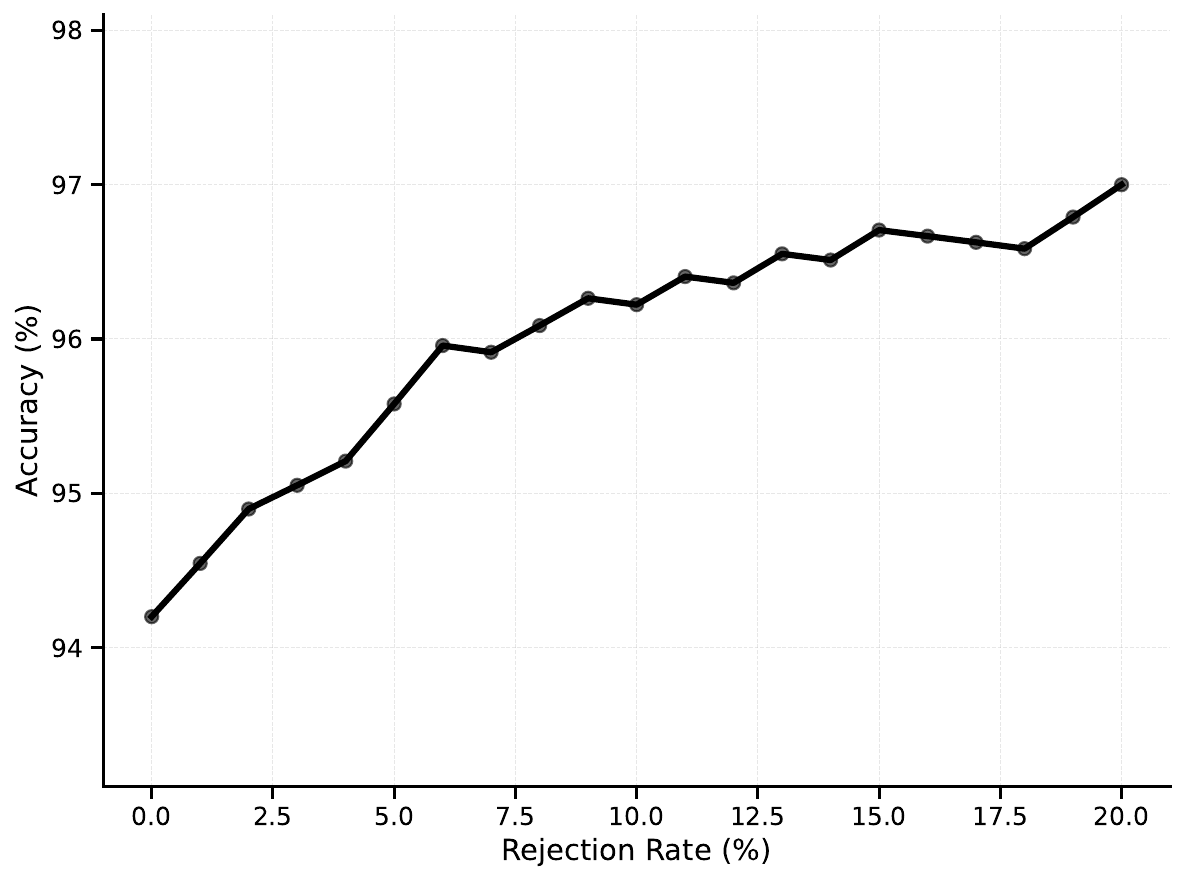}
        \caption{DeepSeek R1}
        \label{fig:selective_prediction_b}
    \end{subfigure}
    \hfill
    \begin{subfigure}[b]{0.3\textwidth}
        \centering
        \includegraphics[width=\textwidth]{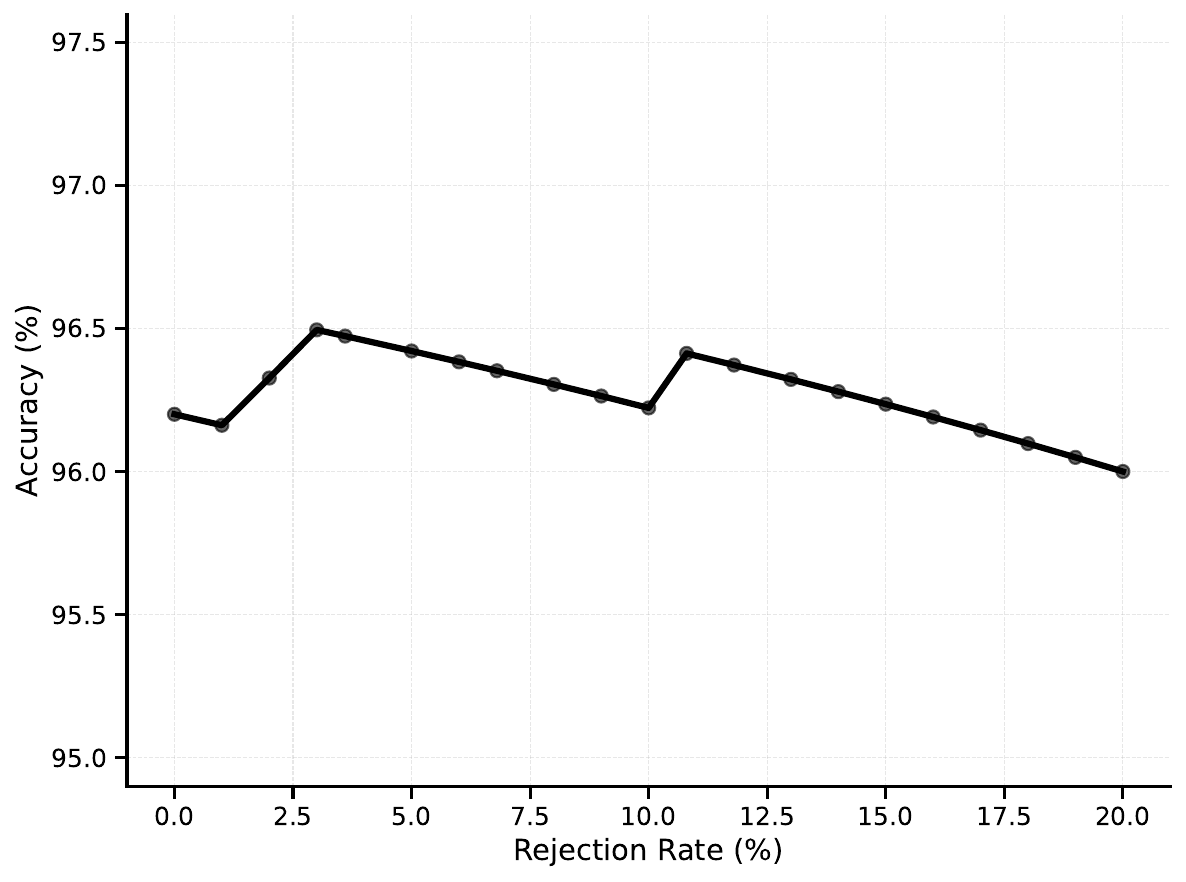}
        \caption{OpenAI o3}
        \label{fig:selective_prediction_c}
    \end{subfigure}
    \caption{Abstaining from answering difficult queries based on the length of the reasoning trace yields 99+\% accuracy for Qwen3 235B-A22B and 97\% accuracy for DeepSeek R1, but yields no performance gains for OpenAI o3.}
    \label{fig:selective_prediction}
\end{figure}

Since the length of reasoning models' thinking traces correlates with decreased accuracy, queries leading to long reasoning traces carry a high risk of error. How much can we raise a model's accuracy by deferring such ``risky'' queries to a human expert? Figure \ref{fig:selective_prediction} shows curves of the conditional accuracy
\begin{equation}
    \text{Accuracy} = \mathbb{E}[\text{Correct~} | \text{~\# Output Tokens} \leq T]
\end{equation}
for different thresholds $T$ of the number of output tokens. We consider thresholds yielding rejection rates from 0\% to 20\% ($x$ axis).

The figure reveals that both Qwen3 235B-A22B and DeepSeek R1 yield increased accuracy when avoiding ``risky'' queries leading to long reasoning traces. For the Qwen3 model, deferring 7.5\% of queries reduces the error rate to 1\% from 3\%—a 2\% reduction. For DeepSeek R1, accuracy gradually improves from 94\% to 97\% when deferring up to 20\% of queries, although the majority of the gains (+2\% accuracy) are already attained for a lower rejection rate of 7.5\%.

By contrast, OpenAI o3 does not yield increased accuracy from avoiding ``risky'' queries with long reasoning traces. We speculate that the diminished efficacy of thinking duration as a proxy for uncertainty stems from OpenAI's proprietary optimization of the thinking process, compared to the simple reinforcement learning recipes described in the technical reports for the open-source models (\citealp{deepseekai2025}; \citealp{yang2025}). In the absence of other techniques for quantifying the uncertainty of o3, this finding places o3 at a disadvantage relative to other reasoning models.

\subsection{What latency and cost savings does ``Failing Fast'' yield?}

\begin{figure}[htbp]
    \centering
    \begin{subfigure}[b]{0.3\textwidth}
        \centering
        \includegraphics[width=\textwidth]{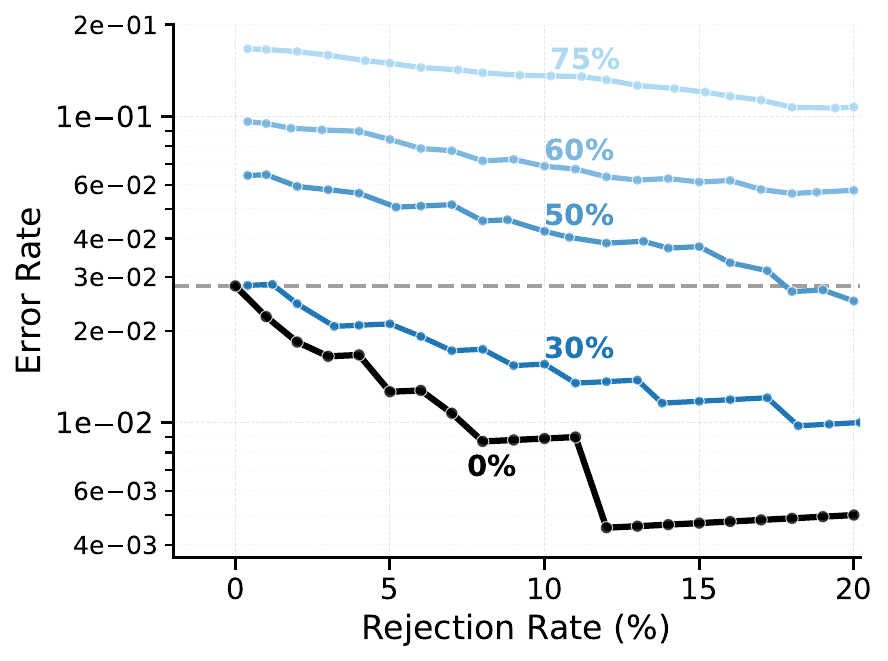}
        \caption{Qwen3 235B-A22B}
        \label{fig:utilization_a}
    \end{subfigure}
    \hfill
    \begin{subfigure}[b]{0.3\textwidth}
        \centering
        \includegraphics[width=\textwidth]{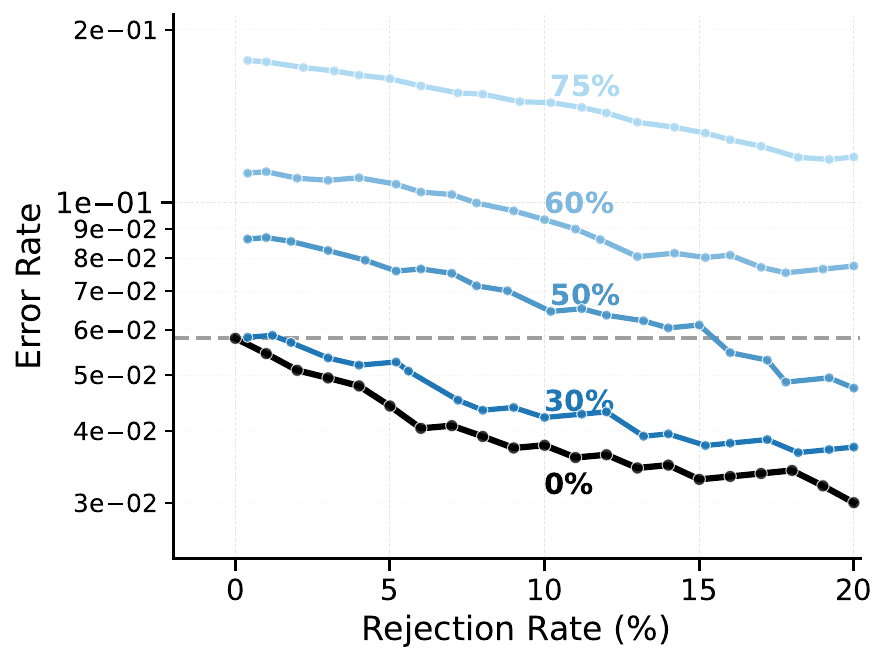}
        \caption{DeepSeek R1}
        \label{fig:utilization_b}
    \end{subfigure}
    \hfill
    \begin{subfigure}[b]{0.3\textwidth}
        \centering
        \includegraphics[width=\textwidth]{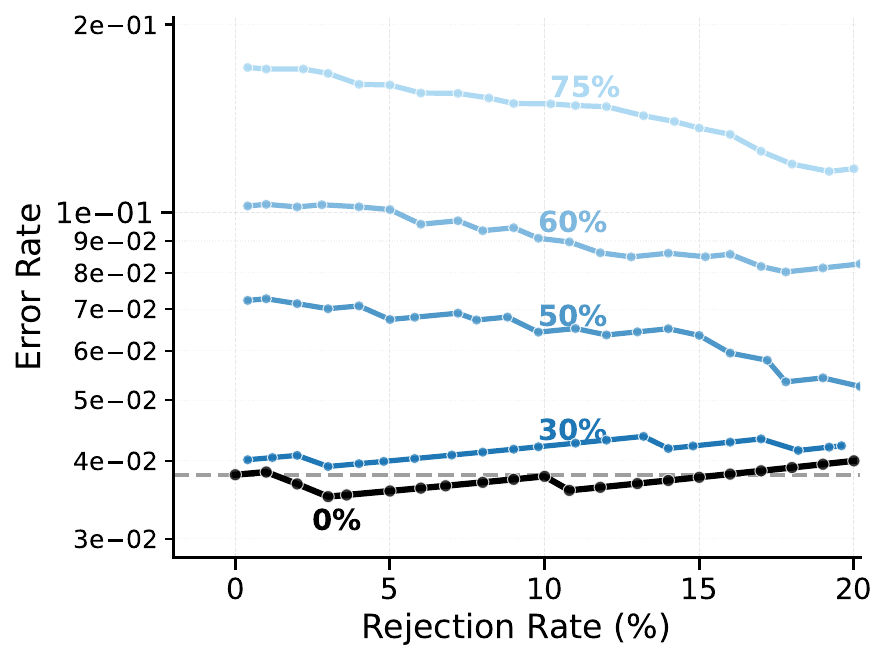}
        \caption{OpenAI o3}
        \label{fig:utilization_c}
    \end{subfigure}
    \caption{Different degrees of non-reasoning model utilization (\% values annotated in color) yield parallel error curves that decrease exponentially with increasing rejection rate ($y$ axis is log-scale). 0\% indicates using only the reasoning model. The dashed gray line indicates the reasoning model's baseline error rate.}
    \label{fig:utilization}
\end{figure}

Deferring risky queries from the reasoning model $\mathcal{M}_r$ to a human expert reduces AI errors. However, it does not address the high latency of reasoning models, which is problematic for interactive use cases or batch workloads with a large volume of queries. Figure \ref{fig:utilization} shows the error-rejection trade-offs when fronting the reasoning model $\mathcal{M}_r$ by a faster non-reasoning model $\mathcal{M}_{nr}$, as described in Section \ref{sec:core}.

The figure shows that when using Qwen3 235B-A22B or DeepSeek R1 as the reasoning model, processing 30\% to 75\% of queries with Llama3.1 405B yields exponentially decreasing error for rejection rates from 0\% to 20\%. However, error decreases faster when using only the reasoning model (indicated as 0\% in the figure, cf. Figure \ref{fig:selective_prediction}). Notably, 50\% utilization of Llama3.1 405B with 15-20\% deferral suffices to push the systems' error rates below the reasoning models' baseline error rates.


Table \ref{tab:savings_table} quantifies the latency and cost savings resulting from 50\%, 60\%, and 75\% utilization of the non-reasoning model. To measure performance holistically across rejection rates, we report the area under the accuracy-rejection curve (AUARC), that is, the mean accuracy across rejection rates from 0\% to 20\%. The table shows that 50\% utilization of Llama3.1 405B incurs a 3\% drop in AUARC, but cuts latency by $\approx27\%$  (Qwen3 235B-A22B and DeepSeek R1), and cost by $\approx 36\%$ (DeepSeek R1). Raising utilization to 60\% yields higher latency reductions of $\approx 40\%$ while still maintaining AUARC above 90\%—significantly outperforming Llama3.1 405B's baseline accuracy of $\approx69\%$.

Finally, 75\% utilization of Llama3.1 405B reduces AUARC to the mid-80\% range but yields correspondingly larger reductions in latency ($\approx55\%$ for Qwen3 and DeepSeek R1) and cost (64\% for DeepSeek R1).

\begin{table*}[t]
\centering
\small
\caption{Human-in-the-loop reasoning model fronted by non-reasoning model maintains high selective prediction performance (90\%+ AUARC $\uparrow$) while saving up to 38\% latency and 46\% cost. $\% \Delta$ denotes performance drop from base AUARC as percentages. $\Delta$L(\%) and $\Delta$C(\%) denote percentage drops in latency and cost.}
\label{tab:savings_table}
\resizebox{\textwidth}{!}{%
\begin{tabular}{lccccccccccccc}
\toprule
\multirow{2}{*}{\textbf{Model}} & \textbf{Base} & \multicolumn{4}{c}{\textbf{50\% Non-Reasoning}} & \multicolumn{4}{c}{\textbf{60\% Non-Reasoning}} & \multicolumn{4}{c}{\textbf{75\% Non-Reasoning}} \\
\cmidrule(lr){2-2} \cmidrule(lr){3-6} \cmidrule(lr){7-10} \cmidrule(lr){11-14}
& AUARC & AUARC & $\Delta$ & $\Delta$L(\%) & $\Delta$C(\%) & AUARC & $\Delta$ & $\Delta$L(\%) & $\Delta$C(\%) & AUARC & $\Delta$ & $\Delta$L(\%) & $\Delta$C(\%) \\
\midrule
deepseek-r1-0528 & 0.96 & 0.93 & \textbf{-2.8} & -26.2 & \textbf{-35.7} & 0.91 & \textbf{-5.4} & -36.5 & \textbf{-45.8} & 0.85 & \textbf{-11.3} & -54.8 & \textbf{-64.1} \\
qwen3-235b-a22b & \textbf{0.99} & \textbf{0.96} & -3.4 & \textbf{-28.4} & -2.1 & \textbf{0.93} & -6.2 & \textbf{-37.8} & -11.6 & \textbf{0.86} & -12.6 & \textbf{-55.7} & -29.7 \\
o3-2025-04-16 & 0.96 & 0.94 & \textbf{-2.8} & -13.4 & -14.8 & 0.91 & -5.5 & -23.2 & -24.9 & 0.85 & -11.4 & -47.5 & -44.4 \\
\bottomrule
\end{tabular}%
}
\end{table*}

\section{Discussion}
\label{sec:discussion}

To mitigate the practical deficiencies of a reasoning LLMs—imperfect accuracy and high latency—we have embedded a reasoning model $\mathcal{M}_r$ into a larger system including a fast non-reasoning model $\mathcal{M}_{nr}$ and a human expert $\mathcal{H}$. As figures \ref{fig:utilization_a} and \ref{fig:utilization_b} show, this system is capable of beating the reasoning model's accuracy while cutting the latency by around 30\%.

However, our methodology treats the human expert as an omniscient oracle. One way to model $\mathcal{H}$ is \textit{learning to defer} (\citealp{madras2018}). However, this approach adapts the AI model to the human, which seems impractical for API-based usage of large reasoning LLMs. Instead of attempting to adapt a trillion-parameter neural network that cost millions of dollars to train, it appears more sensible to adapt the human expert to the LLM.

For example, suppose the human expert $\mathcal{H}$ makes mistakes at a rate $\mathbb{E}[\mathds{1}_\text{error}^\mathcal{H}]$. It is clear that we should select $\mathcal{H}$ to minimize this error rate. However, there is an important nuance. It is critical to minimize the error rate over the \textit{difficult queries},
\begin{equation}
    \mathbb{E}_{q \sim p_\text{difficult}}[\mathds{1}_\text{error}^\mathcal{H}(q)],
\end{equation}
where $p_\text{difficult}(\cdot)$ samples queries for which the reasoning model has a high error rate. In other words, we wish to avoid an ``accuracy drag'' whereby a correlation between the AI model's and the human's perceptions of difficulty raises the error rate of $\mathcal{H}$ conditioned on the deferral decision $\mathcal{M}_r \rightarrow \mathcal{H}$.

This observation has practical implications. For example, suppose we wish to construct a human-in-the-loop system for solving challenging mathematics problems. Among the pool of possible mathematics experts, many of the most able mathematicians likely underperform the LLM on mechanical or otherwise unstimulating questions, simply because they find such questions boring. However, they would likely outperform on ``interesting'' questions whose solutions require inspiration and imagination—the same problems on which AI models are most likely to make mistakes (we speculate). Hence, a screening exam for identifying suitable mathematics experts should only include ``interesting'' questions whose solutions require ingenuity; adding any easier problems would be counterproductive.

\section{Conclusion}
This paper explored reducing the error rate and latency of a reasoning model by embedding it within a larger system that includes a human expert.

First, our experiments show that deferring difficult queries to a human expert can yield meaningful error reductions. Leveraging selective prediction based on the length of the thinking trace, we lowered the conditional error rate of Qwen3 235B-A22B and DeepSeek R1 by around 2\% on difficult MATH questions—from 2.8\% to 0.5\% and 5.8\% to 3\%, respectively. Second, we found that fronting a reasoning model with a large non-reasoning model yields substantial (\~30\%) latency reductions while outperforming the reasoning model's baseline accuracy. Unfortunately, these latency savings are lower than expected because of \textit{latency drag}—the phenomenon that conditioning on difficult queries shifts the reasoning model towards longer latencies than usual. 

Overall, our work offers a novel perspective for adapting reasoning LLMs to the demands of practical use cases. In future work, we are interested in circumventing latency drag by quantifying the uncertainty of a reasoning model in a manner uncorrelated with its latency. In addition, we aim to explicitly model the capabilities of human experts.

\bibliography{main}

\end{document}